\documentclass{article}

     \usepackage[preprint]{neurips_2021}

\usepackage[utf8]{inputenc} 
\usepackage[T1]{fontenc}    
\usepackage[hidelinks,backref=page]{hyperref}       
\usepackage{url}            
\usepackage{booktabs}       
\usepackage{amsfonts}       
\usepackage{nicefrac}       
\usepackage{microtype}      
\usepackage{xcolor}         
\usepackage{amsmath}
\usepackage{graphicx}

\usepackage{tikz}
\usepackage{float}
\usepackage{comment}
\usepackage{amsmath,amssymb} 
\usepackage{color, colortbl}
\usepackage{caption}
\usepackage{algorithm}
\usepackage{algpseudocode}
\usepackage{bbm}
\usepackage{subcaption}
\usepackage{csquotes}

\renewcommand{\b}[1]{\ensuremath{\mathbf{#1}}}	

\newtheorem{definition}{Definition}[section]
\def \Lw {{\mathcal{L}\b{w}}} 
\def \A {{\mathcal{A}}} 
\def \Aw {{\mathcal{A}\b{w}}}

\usepackage[accsupp]{axessibility}  

\title{Robust Graph Neural Networks using Weighted Graph Laplacian}

\author{%
  Bharat Runwal \\
  Indian Institute of Technology, Delhi\\
  \texttt{bharatrunwal@gmail.com}
  \AND
  Vivek  \\
  Samsung Research Bangalore \\
  \texttt{vivekd010@gmail.com}
  \And
  Sandeep Kumar \\
  Department of Electrical Engineering, IIT Delhi, India \\
  \texttt{ksandeep@iitd.ac.in}
}

\begin{document}

\maketitle

\begin{abstract}
Graph neural network (GNN) is achieving remarkable
performances in a variety of application domains. However,
GNN is vulnerable to noise and adversarial attacks in input data.
Making GNN robust against noises and adversarial attacks is an
important problem. The existing defense methods for GNNs are
computationally demanding and are not scalable. In this paper, we propose a generic framework for
robustifying GNN known as Weighted Laplacian GNN (RWL-GNN).
The method combines Weighted Graph Laplacian learning
with the GNN implementation. The proposed method benefits
from the positive semi-definiteness property of Laplacian matrix,
feature smoothness, and latent features via formulating a unified
optimization framework, which ensures the adversarial/noisy edges
are discarded and connections in the graph are appropriately
weighted. For demonstration, the experiments are conducted
with Graph convolutional neural network(GCNN) architecture,
however, the proposed framework is easily amenable to any
existing GNN architecture. The simulation results with benchmark
dataset establish the efficacy of the proposed method, both in accuracy and computational
efficiency. Code can be accessed at \url{https://github.com/Bharat-Runwal/RWL-GNN}.
\end{abstract}

\section{Introdution}
\label{submission}

\begin{figure}[t]
\centerline{\includegraphics[scale=0.50]{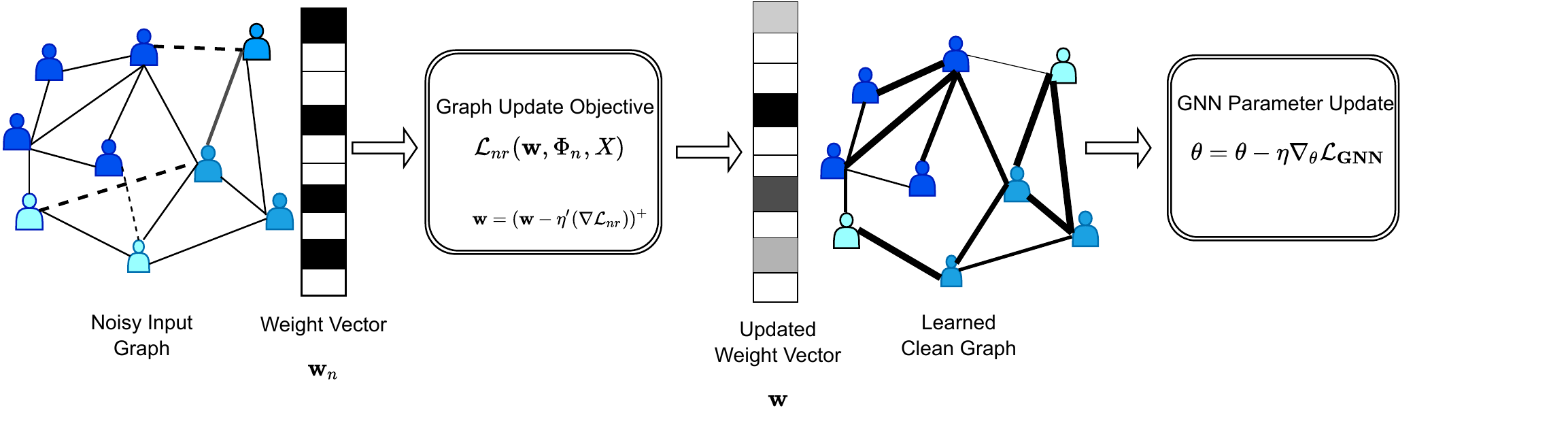}}
\caption{Two-Stage Framework, Dashed Line indicates Adversarial Edges}
\vspace{-1mm}
\label{fig:1-Two}
\end{figure}

\begin{figure}[t]
\hspace*{-0.8cm}
\centerline{\includegraphics[scale=0.50]{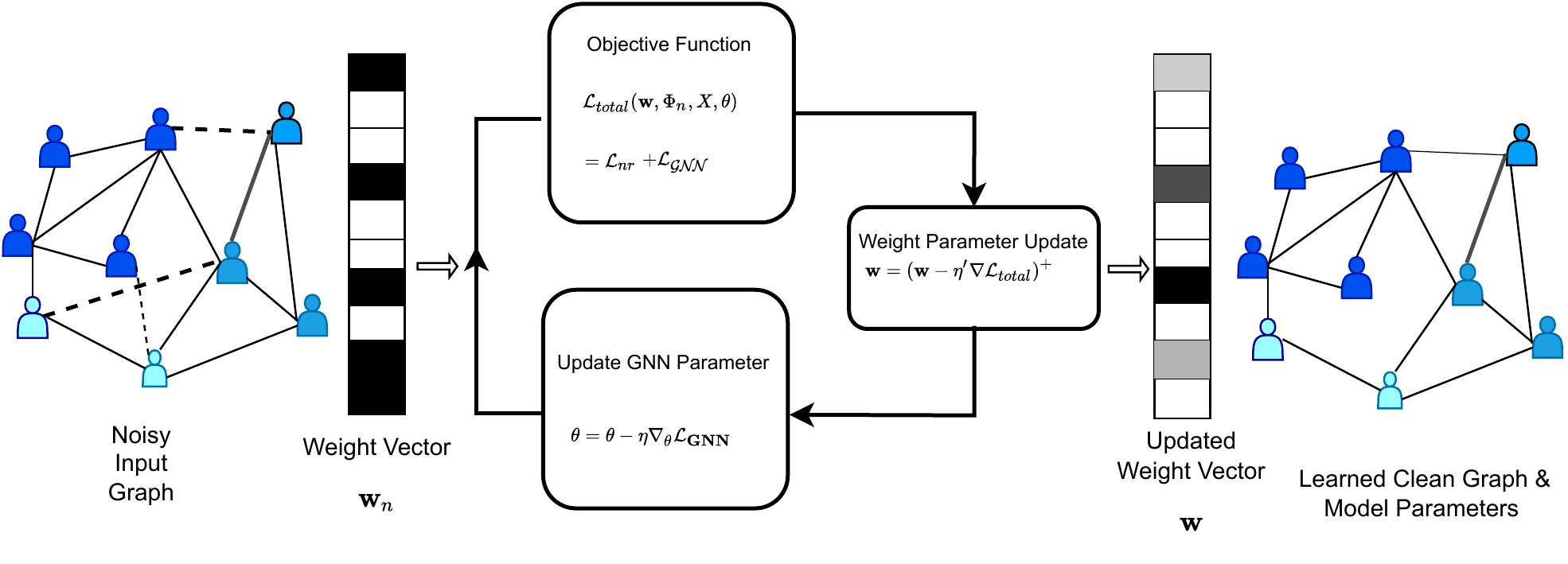}}
\caption{Joint Framework, Dashed Line indicates Adversarial Edges}
\vspace{-1mm}
\label{fig:2-Joint}
\end{figure}

Graphs are fundamental mathematical structures consisting of a set of nodes and weighted
edges connecting them. The weight associated with each edge represents the similarity between the two connected nodes (Homophily) \cite{doi:10.1146/annurev.soc.27.1.415}. We encounter Graph data structure in numerous
domains such as computational networks in social sciences,
financial networks, brain imaging networks, networks in genetics and proteins \cite{Yue2020GraphEO,Ashoor2020GraphEA,Zhang2022DeepLO}. Graphs models the rich relationships between different entities, so
it is crucial to learn the representations of the graphs.
Graph neural network (GNN), a popular deep learning framework for graph data is achieving remarkable performances in a variety of such application domains. GNNs employs a message passing scheme \cite{gilmer2017neural} where the information from the neighborhood of
a particular node is aggregated, transformed and is used for learning the node embeddings. These nodes embedding capture
the structural and the feature-based information of the graph data \cite{Graph_representation}.
\\
However, the question arises is that: are these models reliable. This question on reliability exists due to the existence of
adversarial examples, which are carefully crafted instances
that can occur during the train (Poisoning) or test (Evasion) phase and can fool our
model into making wrong predictions \cite{article}. Even very small deliberate perturbations in the graph can lead to wrong classification or prediction, and these perturbations we call adversarial examples.
These adversarial examples question the
reliability and robustness of these networks in safety-critical
tasks, for example in financial systems \cite{10.1145/3447548.3467145}, medical domain \cite{Finlayson2018AdversarialAA}, etc. Attack methods can be categorized by the attacker's goal, capacity, perturbation type, and knowledge \cite{jin2020adversarial}.
There exists a wide spectrum of
attacks methods on graphs either by perturbing the graph structure or injecting noise into the node features, but most adversarial attacks are done on the graph structure by adding/removing/rewiring edges \cite{jin2020adversarial}. Also, there are certain empirical observations that
can be made from previous works, the first observation is that the attacker prefers
adding edges to perturb the graph instead of removing edges  \cite{jin2020graph}
and second, the attacker tries to connect nodes with dissimilar features, as it disrupts the feature smoothness property \cite{McPherson2001BirdsOA}. In this work, we focused on defending against the adversarial attacks which perturb the graph structure only. 
\\
There exists a wide spectrum of methods that can be used to defend against these
attacks and these methods have been categorized into different categories such as Adversarial training \cite{Dai2018AdversarialAO}, graph purification \cite{10.1145/3336191.3371789,jin2020graph,Zhang2020DefenseVGAEDA}, Attention mechanism \cite{schlichtkrull2017modeling,zhang2020gnnguard}, adversarial perturbation detection \cite{Xu2018CharacterizingME,Ioannidis2019GraphSACDA} etc. \cite{jin2020adversarial}. It has been shown that the attacks connecting nodes with dissimilar features enlarge the singular values \cite{jin2020graph} and also the targeted attack methods try to
perturb the small singular values in adjacency matrix \cite{10.1145/3336191.3371789}, so following these findings they \cite{10.1145/3336191.3371789} proposed a method which uses truncated SVD formulation to get
the low-rank approximation. But these methods may also
remove normal edges in the process of graph purification,
so the alternative purification strategy is to use the graph
learning approach, which aims to learn the graph structure by removing adversarial edges while leveraging the characteristics of adversarial attacks in the process of guiding to
learn the graph structure. Based on empirical observations it is seen that, while attacking the graph models, some of the properties like feature smoothness, low rank, and sparsity are violated \cite{jin2020graph}. 
\\
In this work, we propose a generic framework namely, Weighted Laplacian GNN (RWL-GNN) for robustifying graph neural network architectures against poisoning adversarial attacks. The method is based on obtaining a clean and weighted graph representation in form of a Weighted Laplacian matrix by solving an optimization problem using feature smoothness and the positive semi-definite property of the Graph Laplacian matrix. We further derived two different algorithmic implementations of this framework, i) A two-stage approach, see Fig \ref{fig:1-Two} which first preprocess the given perturbed/noisy graph and obtains a clean graph, which is followed by learning GNN model parameters, and ii) a joint approach, see Fig \ref{fig:2-Joint}, which cleans the noisy Graph Laplacian matrix and learns the GNN model parameters jointly. Please see Fig \ref{fig:1-Two} and Fig \ref{fig:2-Joint} for schematic details of the two proposed methods.

\section{Related Work}

\textbf{Graph Neural Networks.} The defining feature of a GNN is that it uses neural message passing in which messages are exchanged
between nodes and updated using neural networks \citep{gilmer2017neural}. The graph convolution network(GCN) \citep{kipf2017semisupervised}, employs the symmetric-normalized aggregation as well as
the self-loop update approach. The graph convolutions here
are defined as aggregation and transformation of the local
information. The most basic neighborhood aggregation operation simply takes the sum of the neighbor embedding.
Other aggregation fuctions have also been defined namely
the mean, LSTM and Pooling aggregators
etc \cite{10.5555/3294771.3294869}. 
There are other popular GNNs which are proposed recently \cite{velickovic2018graph,Xu2018RepresentationLO,Zeng2020GraphSAINTGS}.

\textbf{Background on Adversarial Attacks.} Traditional DNNs are vulnerable to carefully designed adversarial attacks and GNNs are no exceptions \cite{Dai2018AdversarialAO,Zgner2018AdversarialAO,Zugner2019AdversarialAO}. These attacks are designed in such a way that the victimized nodes
can’t be noticed easily thus producing wrong results. The spectrum of attack methods based on the attacker's capacity can be divided into two categories, Poisoning attack, where the attacker perturbs the training graph, and Evasion attack where the attacker perturbs the graph in test time. According to the attacker's goal, the poisoning attacks can be divided into the targeted attack, where the attacker aims to poison the subset of nodes which led a trained model to misclassify on these targeted nodes, and non-targeted attacks, where the attacker aims to poison all the nodes, in order to degrade the trained model performance on all test data. The targeted attack can be further classified into the direct targeted attack, where the attacker perturbs the edges/features of the nodes directly, and influencer attack, where the attacker doesn't perturb the nodes directly but influences the other nodes to do so \cite{jin2020adversarial}.
Nettack \cite{Zgner2018AdversarialAO} is a targeted attack, which generates the structure and feature attacks which preserve the degree distribution and constraints the feature co-occurrence so that the degradation of GNN model's performance on the downstream task is maximum. The metattack \cite{Zugner2019AdversarialAO} is an untargeted attack that generates attacks based on the meta-learning.
The RL-S2V method is an evasion attack
using reinforcement learning\citep{10.1145/3411501.3419424}. Earlier works have limitations for attacks on large graphs, recently \cite{Geisler2021RobustnessOG} proposed two sparsity-aware first-order optimization attacks for large-scale graphs. 

\textbf{Background on Defense Methods.} There is a growing interest in increasing the robustness of the models to adversarial attacks on graphs. 
Adversarial training, GraphAT \citep{feng2019graph} incorporates node features-based adversarial samples into the training procedure of the classification model. There exist graph purification methods which can be divided into two 1) Pre-Processing based 2) Graph Learning. GNN-Jaccard \cite{Wu2019TheVO} proposed to use a pre-processing stage that removes the edges which have small Jaccard-similarity, this was based on the empirical observation that attacker tends to connect the dissimilar nodes. The Pro-GNN \citep{jin2020graph} formulation is
defined based on the properties of the graph that are affected the
most from the poisoning of graph structure which includes sparsity, feature smoothness, and low rank, but their framework suffers from the heavy computations of eigendecomposition. To increase the robustness of the GNN model itself, the Relational-GCN \citep{schlichtkrull2017modeling} tries to model the hidden representation with Gaussian Distribution, as it can
absorb the information due to the perturbation and leads to
a more robust hidden representation. The PA-GNN  \cite{Tang2020TransferringRF} works on
attention-based mechanism also, but it tries to transfer knowledge from clean graphs where the adversarial attacks can be
generated to serve as supervision signals to learn the desired
attention scores. The GNNGuard \citep{zhang2020gnnguard} tries to quantify a relationship between the graph structure and node features, if one exists, and then exploit that relationship to mitigate the effects of the perturbations. This approach assigns attention scores based on the hidden layer representation and then tries to prune the edges using a user-defined threshold.

\section{Problem Formulation and Background}

A graph is defined by the triplet $\mathcal{G} =(\mathcal{V},\mathcal{E},W)$, where $\mathcal{V}$ is the vertex set containing $n$ nodes $\{v_1,v_2,\dots,v_n \}$, $\mathcal{E} \subseteq \mathcal{V}\times \mathcal{V}$ is the unordered edge set of all possible combinations of $\{(i,j)\}^{n}_{i,j = 1}$ and ${W} \in \mathbb{R}^{n\times n}$ is the adjacency(weight) matrix. We consider only undirected positively weighted graph, with no-self loop, i.e., $ W_{\text{ij}} \geq 0\;\forall\; i\neq j$ and $W_{\text{ii}}=0\;\forall\; i=1,2,\dots,n$.
Graphs can be conveniently captured by some matrix representation, e.g., Laplacian and Adjacency Graph matrices, whose entries correspond to edges in the graph. The choice of a matrix usually depends on modeling assumptions, properties of the desired graph, applications, and theoretical requirements. A Graph Laplacian matrix $\Phi \in \mathbb{R}^{n \times n}$ belongs to the following set: 
\begin{align}\label{constraint}
\mathcal{C}_L =\big\{ \Phi=\Phi^\top\vert \Phi_{ij} \leq 0 \ {\rm for} \ i\neq j; \Phi_{ii}=-\sum_{j\neq i}\Phi_{ij} \big\}. 
\end{align}

\noindent The adjacency matrix $W$ and Laplacian matrix $\Phi$ associated with the graph are related as follows:
$$W_{ij}=\begin{cases}
-\Phi_{ij} \; \text{for}\; i\neq j\\
0\;\;\; \;\;\;\text{for}\; i=j
\end{cases}
$$
Both $\Phi$ and $W$ represent the same graph, however, they have very different mathematical properties. By construction, a Laplacian matrix $\Phi$ is positive semidefinite, implied from the diagonally dominant property ($|\Phi_{ii}|\geq \sum_{j\neq i, }^n|\Phi_{ij}|\; \forall \; i=1,2,\ldots,n $) and a $M$-matrix, i.e., a positive semidefinite matrix with non-positive off-diagonal elements \cite{slawski2015estimation}. In addition, a Laplacian matrix has zero row sum and column sum, i.e., $\Phi_{ii}+\sum_{i\neq j}\Phi_{ij}=0$ which means that the vector $\boldsymbol{1} = [1, 1, \ldots, 1]$ satisfies $\Phi\boldsymbol{1} = \boldsymbol{0}$ \cite{ kumar2020unified}. Owing to these properties, Laplacian matrix representation is more desirable for building graph-based algorithms.

The feature matrix is denoted by $X = [x_1,x_2,\dots,x_n] \in \mathbb{R}^{n \times d}$, where d is the dimensions of each features and $x_i$ is the feature vector of node $v_i$. In the semi-supervised node classification setting we will train on given subset of nodes $\mathcal{V}_{L} = \{v_{1},v_{2},\dots,v_{l}\} $ with the corresponding labels
 $\mathcal{Y}_{L}=\{y_{1},y_{2},\dots,y_{l}\}$. The goal of training is to learn the function $f_{\theta}$, i.e., $f_{\theta}: \mathcal{V}_{l} \rightarrow \mathcal{Y}_{l}$ which can classify the unlabeled nodes to the correct classes function  \cite{kipf2017semisupervised}. The training objective function can be formulated as: 
 \begin{equation}\label{gnn-obj}
\mathcal{L_{\textit{GNN}}}(\theta,\Phi,X,y_{l})= \sum_{\textit{u}_{i} \in \mathcal{V_{L}}}\textit{l}(f_{\theta}(X,\Phi)_{i},y_{i})
\end{equation}
where $f_{\theta}(X,\Phi)_{i}$ and $y_{i}$ are the predicted and the true label of node $v_i$, $\theta$ is the learnable parameters of the GNN model, $\Phi$ is the Graph Laplacian matrix, and $\textit{l}(.,.)$ is a loss function such as cross-entropy. However, in an adversarial scenario, the graph information is noisy/perturbed, which means we have a noisy version of the graph Laplacian matrix, denoted as $\Phi_{\text{n}}$. Training the GNN model \eqref{gnn-obj} with noisy graph information will not yield a reliable predictor function $f_\theta$. Thus we propose a framework, which trains the GNN model after removing the noise from the graph matrix, we aim to solve the following optimization problem:
\begin{subequations}\label{gnn-joint} 
\begin{equation}
\min_{\theta}\;\; \mathcal{L_{\textit{GNN}}}(\theta,\Phi^*,X,y_{l}) \label{gnn-joint-1} 
\end{equation}
\begin{equation}\text{and}\;\;\Phi^*=\arg\min_{\Phi}\mathcal{L_{\text{nr}}}(\Phi_n,X,\Phi) \label{gnn-joint-2}
\end{equation}
\end{subequations}
where GNN model is trained on a clean graph Laplacian $\Phi^*$, which is obtained by minimizing a noise removal objective function: $\mathcal{L_{\text{nr}}}(\Phi_n,X,\Phi)$. In the next section, we discuss two methods for solving the proposed problem \eqref{gnn-joint} tractably.

\section{Proposed Framework}
In order to solve \eqref{gnn-joint} tractably, we proposed a joint optimization formulation for learning robust GNN model parameters and clean graph structure from the input data as the following: 
\begin{equation}\label{Optimization problem}
 \underset{\theta,\;\Phi \in \mathcal{C}_{L} }{\text{minimize}} \;\; \mathcal{L_{\textit{total}}} = \mathcal{L_{\textit{GNN}}}(\theta,\Phi,X,y_{l})+\mathcal{L_{\text{nr}}}(\Phi,\Phi_{n},X)
\end{equation}
where $\Phi_{n}$ and $X$ are noisy input graph Laplacian matrix and feature attribute matrix respectively, $\mathcal{C}_L$ is the Laplacian matrix structural constraint set \eqref{constraint}, $\Phi$ is the clean weighted Graph Laplacian matrix which is to be learned. Finally, the noise removal objective function is defined as:
\begin{align}
 & \mathcal{L_{\text{nr}}}(\Phi,\Phi_{n},X) =\mathcal{\alpha} ||\Phi-\Phi_{n}||_{F}^{2} +\mathcal{\beta} Tr(X^{T}\Phi X)\label{Smoothness objective}
\end{align}
where $\alpha,\beta\geq 0$ are the respective weights. 

\noindent It is remarked that the adversarial attacks focus on introducing unnoticeable perturbations and connecting two nodes with dissimilar features \cite{McPherson2001BirdsOA}. The formulation in \eqref{Smoothness objective} is a potential solution for providing robustness against such types of attacks. More concretely, the term $||\Phi-\Phi_{n}||_{F}^{2}$ tries to maintain the structure of the learned graph by ensuring that the learned Laplacian matrix does not deviate too far from the original input matrix. Next, minimizing $ Tr(X ^{T}\Phi X)=-\frac{1}{2}\sum_{i,j}\Phi_{ij}||x_i - x_j ||^2$, minimizes the Dirichlet energy on the graph and conceptually, it promotes smoothness of $X$ by penalizing high frequency components in the graphs, which ensures that the edges introduced between two nodes with dissimilar features should be removed or down-weighted.

\noindent The above problem \eqref{Optimization problem} can be solved by following both a two-stage approach i.e. first preprocessing the graph to get a clean graph structure and then using it to learn the GNN parameters \cite{10.1145/3336191.3371789, Wu2019TheVO} and a joint approach where we solve the two problems together, i.e., the noise removal problem depending on the GNN parameter and vice versa \cite{jin2020graph}. However, there exists a trade-off between the two approaches, no one is an obvious winner: the two-stage approach is computationally efficient but may provide a suboptimal graph for the GNN model 
when compared to the joint approach \cite{jin2020graph}, on the other hand, the joint approach is computationally demanding but performs well across all perturbation rates but provides less robustness at higher perturbations compared to the two-stage according to our experiments. 
In the next section, we develop generic, computationally efficient methods for solving \eqref{Optimization problem}.

\subsection{Two-Stage Optimization Framework}
\noindent In this section, we propose a two-stage approach for solving \eqref{gnn-joint}, i) in the first step we remove the noise from the input Graph, and ii) in the second step we use the clean graph to learn a robust GNN model.\\
\textbf{Stage 1:} Solving for noise removal objective:
\begin{align}
 & \underset{\Phi \in \mathcal{C}_{L} }{\text{minimize}} \; \mathcal{L_{\text{nr}}}(\Phi,\Phi_{n},X) \label{Optimization Term}
\end{align}
This is a Laplacian structural constrained matrix optimization problem where $\Phi\in \mathcal{C}_L$. We simplify the matrix constraints to simple non-negative vector constraints by introducing Laplacian operator defined in \cite{kumar2020unified}, that maps a vector $\b{w} \in \mathbb{R}^ {n(n-1)/2}$ to a matrix ${{\mathcal{L}}} \b{w} \in \mathbb{R}^{n \times n}$ which satisfies the Laplacian constraints ($[{\mathcal{L}} \b{w}]_{ij} =[{\mathcal{L}} \b{w}]_{ji}, {\text{for}} \, i\neq j $ and $[{\mathcal{L}} \b{w}] \mathbf{1} =\mathbf{0}$). 

\begin{definition}
Laplacian operator ${{\mathcal{L}}}$ : $\mathbb{R}^ {n(n-1)/2}\rightarrow \mathbb{R}^{n \times n}$, $\b{w} \mapsto \mathcal{L}\b{w}$ is defined as :
\[ 
[{\mathcal{L}} \b{w}]_{ij}= \left\{
\begin{array}{ll}
-\b{w}_{i+d_j}\; & \; \;i >j,\\
\hspace{.2cm}[\mathcal{L}\b{w}]_{ji} \; & \; \;i<j,\\
-\sum_{i\neq j}[\mathcal{L}\b{w}]_{ij}\; &\;\; i=j,
\end{array} 
\right. 
\]
where $d_j=-j+\frac{j-1}{2}(2n-j).$
\end{definition}
\begin{definition}
The adjoint operator ${{\mathcal{L}^{*}}} :Y \in \mathbb{R}^{n \times n}\rightarrow{{\mathcal{L}^{*}}}Y\in \mathbb{R}^{n(n-1)/2}$ is defined by: 
	\begin{equation*}
	[{\mathcal{L}} ^* Y]_k=Y_{i,i}-Y_{i,j}-Y_{j,i}+Y_{j,j},\; \; k=i-j+\frac{j-1}{2}(2n-j),
	\end{equation*}
		where $k=i-j+\frac{j-1}{2}(2n-j)$ and $i,j \in \mathbb{Z}^+, i>j$.
\end{definition}
The operator ${{\mathcal{L}}}$ and its adjoint ${{\mathcal{L^{*}}}}$ satisfies the following condition $ \langle {{\mathcal{L}}} \b{w},Y \rangle=\langle \b{w},{{\mathcal{L}}}^*Y \rangle$. We can similarly define the adjacency operator $\A$.
Using the Laplacian operator $\mathcal{L}$, we can simplify the Laplacian set $\mathcal{C}_L$ in \eqref{constraint} as below:
\begin{align}\label{lapset-reformulation}
 \mathcal{C}_L=\{\mathcal{L}\b{w}\vert \b{w}\geq 0\}
\end{align}
Replacing $\Phi$ with $\Lw$ and reformulating the constraints as in \eqref{lapset-reformulation}, the optimization problem
 in \eqref{Optimization Term} can now be expressed as: 
\begin{equation}\label{Final Problem}
 \min_{\b{w} \geq 0 } \mathcal{L_{\text{nr}}} = \mathcal{\alpha} ||\Lw-\Phi_{n}||_{F}^{2} 
 +\mathcal{\beta} Tr(X^{T}\Lw X)
\end{equation}
This is a non negative constrained convex quadratic program :
\begin{equation}\label{NNP}
 \min_{\b{w} \geq 0} f(\b{w}) = ||\Lw||_{F}^{2} - \b{c}^{T}\b{w}
\end{equation}
where $\b{c} ={{\mathcal{L}^{*}}}[\Phi_{n} - \frac{\beta}{\alpha} XX^{T}]$.

\textbf{Note:} $\b{c}$ can be precomputed before the training for a given dataset and perturbed graph structure. 

Due to the non-negativity constraint $w \geq 0$, the problem \eqref{NNP} doesn't have a close form solution. In order, to make the algorithm scalable, we employ the majorization-minimization framework \cite{sun2016majorization}, where we obtain easily solvable surrogate functions for objective functions such that the update rule is easily obtained. Thus we perform first-order majorization of $f(\b{w})$ as: 
\begin{align*}
 g\left(\b{w}|\b{w}^{(t)} \right)&=f\left( \b{w}^{(t)}\right) + \left(\b{w} - \b{w}^{(t)} \right)^\top \nabla f \left(\b{w}^{(t)}\right) \\
 &+\frac{L_1}{2} ||{\b{w} - \b{w}^{(t)}}||^2
 \end{align*}
where $L_1 = ||\mathcal{L}||_2^{2}=2n$ Lipschitz constant, see \cite{kumar2020unified} for more details. The new optimization problem is formulated as: 
\begin{equation}
 \min_{\b{w} \geq 0} g(\b{w}|\b{w}^{(t)})= \frac{1}{2}\b{w}^{T}\b{w} - \b{a}^{T}\b{w}
\end{equation}
where $\b{a} = \b{w}^{T} - \frac{1}{L_{1}} \nabla f(\b{w}^{t})$ and $\nabla f(\b{w}^{t})={\mathcal{L}^{*}}{\mathcal{L}}(\b{w}^{t})-\b{c}$. 
Using the KKT Optimality condition we have:
\begin{equation}
 \nabla f(\b{w}^{t}) ={\mathcal{L}^{*}}{\mathcal{L}}(\b{w}^{t})-\b{c}
\end{equation}
Below is our update rule for weights \b{w} is:
\begin{equation}\label{KKT}
 \b{w}^{t+1} = (\b{w}^{t} -\nabla f(\b{w}^{t}))^{+}
\end{equation}
where $x^{+} = \max(0,x)$, $t$ is the iteration step. The iterations are performed until certain stopping criteria are met. The associated adjacency matrix is simply $\mathcal{A}\mathbf{w}$. The weights $\mathbf{w}^*$ correspond to a clean graph, where noisy edges are removed and important connections are appropriately weighed.

\textbf{Stage 2:}
The second stage just learns the GNN parameters using the clean graph adjacency matrix:
\begin{equation}
 \min_{\theta}\mathcal{L_{\textit{GNN}}}(\theta,\Aw,X,y_{l})
\end{equation}
We have described our two-stage optimization framework in the algorithm (\ref{alg:RWL}), where T and T' are the number of epochs respectively for both the stages.
\begin{algorithm}[tb]
 \caption{Proposed Algorithm: Two-Stage}
 \label{alg:RWL}
\begin{algorithmic}
 \State {\textbf{Input}: $\mathcal{G} =(\Phi_{n}, \textbf{X}, y_{l}) $, parameters: $\alpha, \beta$, T, T'} 
 \State {\textbf{Outputs}: $\b{w}$, $\theta$} 
 \State {\bfseries Initialize}: \b{w} $\rightarrow \b{w}_n$, $\theta \rightarrow$ Randomly
 \State {\textbf{Stage 1: Pre-Processing Perturbed Graph}}
 \For{$t=1$ {\bfseries to} $T$}
    \State $ \b{w}^{t+1} = (w^{t} -\nabla f(w^{t}))^{+}$
 \EndFor
 \State {\textbf{Stage 2: GNN Parameter Update }}
 \For{$t=1$ {\bfseries to} $T'$}
 \State $ \theta^{t+1} = (\theta^{t} -\nabla \mathcal{L}_{GNN}(\theta,\mathcal{A}w^{T},X,y_{l}))$
 \EndFor
\end{algorithmic}
\end{algorithm}
\begin{algorithm}[tb]
   \caption{Proposed Algorithm: Joint}
   \label{alg:joint}
\begin{algorithmic}
   \State {\bfseries Input: $\mathcal{G}= (\Phi_{n},\textbf{X} ,y_{l})$,parameters $\alpha,\beta$,T}
   \State {{\bfseries Outputs}: $\b{w},\theta$} 
  \State {{\bfseries Initialize}: w $\rightarrow \text{w}_n$,$\theta \rightarrow$Randomly}
  \While{\textit{Stopping conditions not met} }
  \State {\textbf{Graph weight update}}
    \State $ w^{t+1} = (w^{t} -(\nabla f_{1}(w^{t})+ \nabla f_2(w^{t})))^{+}$

  \State {\textbf{GNN parameter update}}
    \For{$t'=1$ {\bfseries to} $T$}
    \State $ \theta = (\theta -\nabla \mathcal{L}_{GNN}(\theta,\Aw^{T},X,y_{l}))$
  \EndFor

  \EndWhile
  
\end{algorithmic}
\end{algorithm}
\noindent In the next subsection we will describe our second method which jointly cleans the perturbed graph and learns the GNN parameters.

\subsection{Joint optimization Framework}
 Using \eqref{lapset-reformulation} the joint optimization problem in \eqref{Optimization problem} is reformulated as:
\begin{equation}\label{Optimization problem Modified}
 \min_{\theta,\b{w} \geq 0 } \mathcal{L_{\textit{nr}}}(\b{w},\Phi_{n},\text{X}) + \mathcal{L_{\textit{GNN}}}(\theta,\Aw,X,y_{l})
\end{equation}
Collecting the variables as a double ($\b{w},\theta$), and noting that, the constraints are decoupled. We can use block alternating optimization framework for solving this problem, where we solve for each block one at a time, keeping the rest of the blocks fixed. In order, to make the algorithm scalable, we employ the block majorization-minimization \cite{Razaviyayn2013AUC,kumar2020unified}. The block-MM approach is a highly successful framework for handling large-scale and non-convex optimization problems.\\
\noindent Collecting the variables $\mathcal{X}=(\mathbf{w}\in \mathbb{R}^{n(n-1)/2}, \theta)$, we develop a block MM-based algorithm that updates one variable at a time while keeping the other ones fixed. We now describe the update rules for the variables: \\
\noindent \textbf{Update of $\theta$: }
For the update of the model parameters, we fix the weights w and we learn the model parameters by solving the following optimization problem : 
\begin{equation}\label{GNN-obj-joint}
 \min_{\theta} f(\theta)=\mathcal{L_{\text{GNN}}}(\theta,\mathcal{A}\b{w},X,y_{l})
\end{equation}

\noindent \textbf{Note:} In our experiments, We approximate the solution of the above problem by one-step gradient descent.
\begin{equation}
 \theta = \theta - \eta'g
\end{equation}
where $\eta'$ is the learning rate and g is the gradient obtained using PyTorch autograd.
\\
\noindent \textbf{Update of w:}
For the update of weight parameters, we fix the model parameters $\mathcal{\theta}$ and learn the weights by solving the following optimization problem: 
\begin{equation}\label{Optim-w}
 \min_{\b{w} \geq 0} \mathcal{L_{\textit{nr}}}(\b{w},\Phi_{n},X) + \mathcal{L_{\textit{GNN}}}(\theta,\Aw,X,y_{l})
\end{equation}
With the help of the Graph Laplacian and Adjoint operators, we can rewrite the problem \eqref{Optim-w} as : 

\begin{equation}
 \min_{\b{w} \geq 0}f(\b{w}) = f_1(\b{w})+f_2(\b{w})
\end{equation}
where $f_1(\b{w})= ||\mathcal{L{\textit{\b{w}}}}||_{F}^{2} - \b{c}^{T}\text{\b{w}}$, $f_2(\b{w})= \mathcal{L}_{GNN}$, each $\mathcal{\text{\b{w}}}_{i} \geq 0$, $\b{c} ={{\mathcal{L}^{*}}}[\Phi_{n} - \frac{\beta}{\alpha} XX^{T}]$, where $\mathcal{L}^{*}$ is adjoint operator.
The function $f_1(\b{w})$ and $f_2(\b{w})$ can be approximated by the second order Taylor series as following \cite{Gao2021StochasticGN}:
\begin{align*}
g_1\left(\b{w}|{\b{w}}^{(t)} \right)&=f_1\left( {\b{w}}^{(t)}\right) + \left({\b{w}} - {\b{w}}^{(t)} \right)^\top \nabla f_1 \left({\b{w}}^{(t)}\right) \\ 
&+\frac{L_1}{2} ||{{\b{w}} -{\b{w}}^{(t)}}||^2 \\
g_2\left(\b{w}|{\b{w}}^{(t)} \right)&=f_2\left( {\b{w}}^{(t)}\right) + \left({\b{w}} - {\b{w}}^{(t)} \right)^\top \nabla f_2 \left({\b{w}}^{(t)}\right) \\ 
&+\frac{L_2}{2} ||{{\b{w}} -{\b{w}}^{(t)}}||^2
\end{align*}
where $L_1=||\mathcal{L}||_2^{2}=2n$ and $L_2$ are the Lipschitz constants for $g_1$ and $g_2$ respectively. We don't need a tight $L_2$ to satisfy the approximation condition, any $L_2{'} \geq L_2$ will work \cite{kumar2020unified}. The new optimization problem can be formulated as : 
\begin{equation}\label{Majorize-joint}
\min_{\b{w} \geq 0} g_1(\b{w}|\b{w}^{(t)})+g_2(\b{w}|\b{w}^{(t)})
\end{equation} 
Solving the optimization problem \eqref{Majorize-joint}, 
We can get the update rule for the weight parameters as:
\begin{equation}
 \b{w}^{t+1} = \left(\b{w}^{t} -\frac{1}{L_1 + L_2}(\nabla f_{1}(\b{w}^{t})+ \nabla f_2(\b{w}^{t}))\right)^{+}
\end{equation}
where $x^{+} = \max(0,x)$, the gradients $\nabla f_1(\b{w})={\mathcal{L}^{*}}{\mathcal{L}}(\b{w}^{t})-\b{c}$ and we can use the pytorch autograd to get the gradients of $f_2(\b{w})$. 
We summarize our joint optimization algorithm in \ref{alg:joint}.

\textbf{Note} that in algorithm \ref{alg:joint}, in each iteration the graph weights $\b{w}$ are updated once while we can update the model parameters as per our necessity but in most of our experiments except for Random Attack we have updated the model parameters only once, details of this can be found in our experiment section.
\section{Experiments}
In this section, we provide the experimental results of our defense framework against different kinds of adversarial attacks.

\subsection{Setup:}
\noindent We validate our framework on the most commonly used citation network datasets; Cora and Citeseer, the statistics of the datasets used are given in table \ref{Table:Dataset}. We compared our method with current state-of-the-art methods ProGNN \cite{jin2020graph}, GNNGuard \cite{zhang2020gnnguard} and a Two-Stage (preprocessing) method GCN-Jaccard \cite{Wu2019TheVO}.
\begin{table}[H]
\vskip -1em
\centering
\caption{Dataset Statistics. Following \cite{Zgner2018AdversarialAO, Zugner2019AdversarialAO,10.1145/3336191.3371789}, we only consider the largest connected component (LCC).}
\begin{tabular}{c|cccc}
\toprule
    Dataset     & Nodes & Edges  & Classes & Features \\ \midrule
Cora     & 2,485 & 5,069  & 7       & 1,433     \\ 
Citeseer & 2,110 & 3,668  & 6       & 3,703     \\  \bottomrule
\end{tabular}
\label{Table:Dataset}
\end{table}
We followed the same experimental setup as \cite{jin2020graph}. We used the GCN architecture with two layers. For each graph we chose randomly 10\% /10\% /80\% (train/valid/test) of nodes. The hyper-parameter tuning is done based on the validation data. We validate our algorithm on the following three attacks Targeted Attack(Nettack), Non-Targeted attack(Meta-self), and Random Attack:
\begin{itemize}
 \item Targeted Attack: These attacks are done on the given subset of target nodes. We used state-of-the-art targeted attack \textit{nettack} \cite{Zgner2018AdversarialAO} for our experiments. 
 \item Non-Targeted Attack: These attacks downgrade the overall performance of GNNs on the whole graph instead of focussing on subset of target nodes. We used variant(Meta-Self) of representative nontargeted attack \textit{metattack} \cite{Zugner2019AdversarialAO}
 \item Random Attack: In this attack we randomly add noise(fake Edges) to the clean graph structure.
\end{itemize}

We used deeprobust library \cite{li2020deeprobust} for the implementation of attacks and GCN architecture. The attack splits follows the Pro-GNN \cite{jin2020graph} i.e. For Nettack, the nodes in test set with degree greater than 10 are set as target node, and the number of perturbations on every target node is varied from 1 to 5 in the step of 1. For Metattack, the perturbation is varied from 0 to 25\% with steps of 5\%. For Random attack, random noises(addition of edges) from 0\% to 100\% were added with steps of 20\%. We used the stochastic gradient descent(SGD) Optimizer for learning model parameters and weights with learning rate $1e^{-3}$ and learning rate $1e^{-2}$ respectively. For our two-stage framework, in the first stage, we used $\sim200$ Epochs to get the clean graph, and for second stage of learning GNN model parameters, we use early stopping with 200 epochs, and the rest of the settings same as \cite{jin2020graph}. For the Joint optimization framework (\ref{alg:joint}), we have $T=1$ for all of our experiments except for random attack where we used $T=2$. Here also, we used $\sim200$ Epochs. We used $\beta$ according to the perturbation rate, as we increase the perturbation rate we may need a higher value for $\beta$, we used $\beta$ in the range of $0.1-0.5$ and fix $\alpha = 1$, we get the best hyper-parameter values by validating on the validation set. 
\subsection{Results}
\begin{figure}[t!]
{\includegraphics[width=\linewidth]{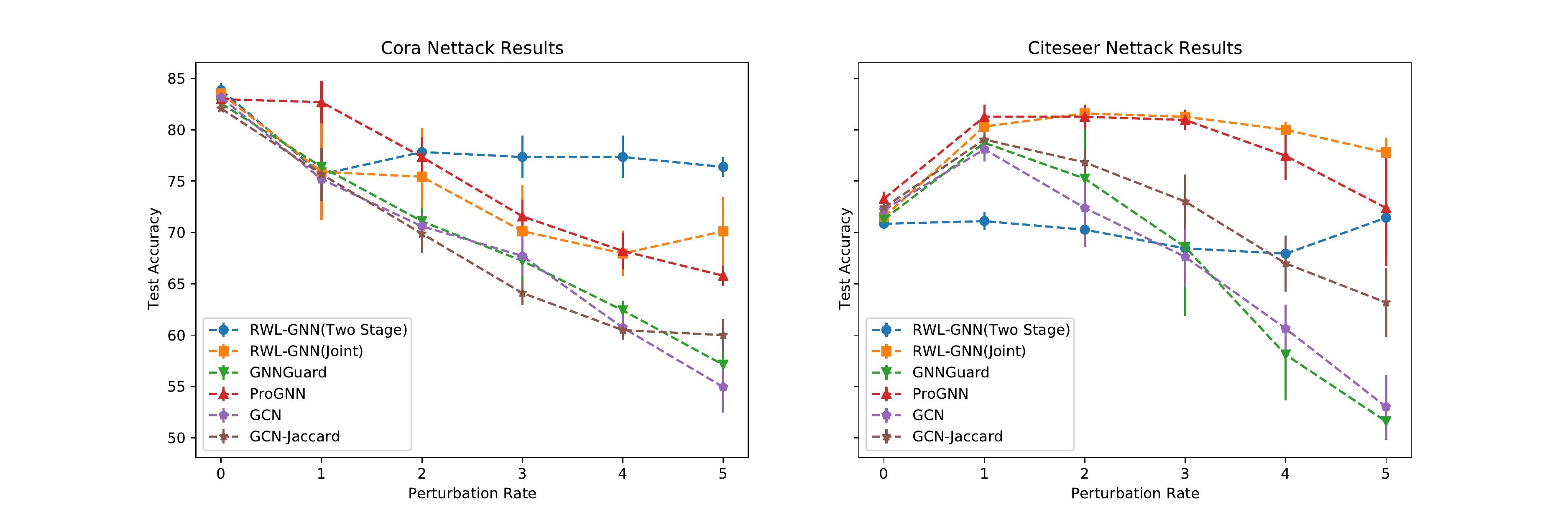}}
\caption{Results of Different models under Nettack}
\label{fig:4}
\end{figure}

\begin{figure}[t!]
{\includegraphics[width=\linewidth]{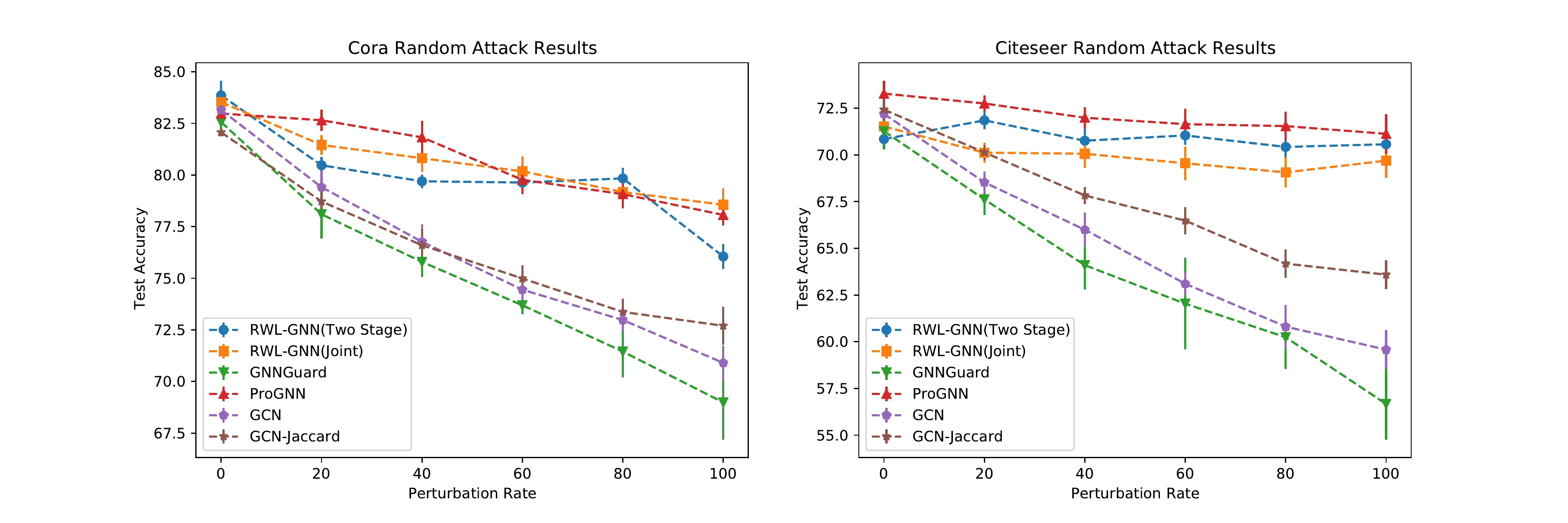}}
\caption{Results of Different models under Random Attack}
\label{fig:5}
\end{figure}

\noindent We provide the node classification accuracy results with average of 10 runs (for Mettack) and 5 runs (for Nettack and Random) for Cora and Citeseer dataset under different attacks for different perturbation rate in Table \ref{tabel1:Meta-attack}, Fig. \ref{fig:4}, \ref{fig:5}. We also show our performance comparison with the two-stage variant of Pro-GNN \cite{jin2020graph} with our proposed two-stage framework in Table \ref{table2:two-stage}. From this, we can clearly see that our two-stage framework performs better with a good margin across all perturbation rates. Also, the number of epochs required for reaching the best accuracy is significantly less compared to Pro-GNN, for example, Nettack Pro-GNN(Joint) uses 1000 epochs to train to achieve good robustness but our algorithm(joint) can converge to best accuracy faster with only 200 epochs, similarly, for our two-stage framework, we require 200 epochs for stage 1 preprocessing, thus saving training time and making our algorithm faster.  Note that the proposed frameworks outperform the others especially at higher perturbation by a large margin in most cases but in some cases, it underperforms compared to the other methods.
In our future work, we will try to improve our framework by using different regularizations for learning better graph structure.   
\begin{table}[!t]
\caption{Meta-attack Experimental Results}
\label{tabel1:Meta-attack}
\centering
\setlength{\tabcolsep}{4pt}
\renewcommand{\arraystretch}{0.95}
\small
\begin{tabular}{c|c|c|c|c|c|c|c}
\toprule
{Dataset} & {Ptb Rate($\%$)}  
&{GCN}&{GCN-Jaccard}&{ProGNN}&{GNNGuard}&{RWL-GNN}&{RWL-GNN} 
\\
 & & &&&&(Two-Stage)& (Joint)
\\
\midrule

{Cora} & 0 

& 83.50$\pm$0.44&82.05$\pm$0.51   &82.98 $\pm$0.23  &82.38$\pm$1.23  & 83.62 $\pm$ 0.821& 83.53 $\pm$0.54 \\

& 5 
&  76.55$\pm$0.79& 79.13$\pm$0.59  & 82.27$\pm$0.45  &78.96$\pm$0.55  & 78.813 $\pm$ 0.95& 79.81 $\pm$0.80\\

& 10 
&  70.39$\pm$1.28& 75.16$\pm$0.76  &79.03$\pm$0.59  &72.17$\pm$2.01   &79.015 $\pm$ 0.843& 78.80 $\pm$0.81\\

& 15 
&  65.10$\pm$0.71& 71.03$\pm$0.64  & 76.40$\pm$1.27  &68.71$\pm$1.87  &    78.295 $\pm$ 0.862& 78.01 $\pm$0.72\\

& 20 
&  59.56$\pm$2.72& 65.71$\pm$0.89  & 73.32$\pm$1.56  &58.48$\pm$1.59  &    78.294 $\pm$ 1.051& 77.59 $\pm$0.60\\
& 25 
&  47.53$\pm$1.96& 60.82$\pm$1.08  & 69.72$\pm$1.69 &53.19$\pm$0.79  &    76.533 $\pm$ 0.72& 75.62 $\pm$0.76\\
\midrule
{Citeseer} & 0 

&  71.96$\pm$0.55& 72.10$\pm$0.63  & 73.28$\pm$0.69  &71.15$\pm$0.84  &    70.842 $\pm$ 0.454& 71.52 $\pm$0.66\\

& 5 
&  70.88$\pm$0.62& 70.51$\pm$0.97  & 72.93$\pm$0.57  &70.78$\pm$0.68  & 71.103 $\pm$ 0.865  & 69.82 $\pm$0.54  \\

& 10 
&  67.55$\pm$0.89& 69.54$\pm$0.56  & 72.51$\pm$0.75  &66.04$\pm$0.81  &70.263 $\pm$ 1.015& 69.15$\pm$0.39\\

& 15 
&  64.52$\pm$1.11 & 65.95$\pm$0.94 & 72.03$\pm$1.11  &64.29$\pm$1.29  &    68.453 $\pm$ 1.306& 65.60 $\pm$0.81\\

& 20 
&  62.03 $\pm$3.49& 59.30$\pm$1.40 & 70.02$\pm$2.28 &58.80$\pm$2.12  &    67.927 $\pm$ 1.231& 65.44 $\pm$0.70\\
& 25 
&  56.94$\pm$2.09& 59.89$\pm$1.47  & 68.95$\pm$2.78  &56.07$\pm$2.65  &    71.433 $\pm$ 0.763& 65.32 $\pm$0.74\\

\bottomrule
\end{tabular}
\end{table}


\begin{table}[t!] 
\caption{Mettack results of Pro-GNN-two and RWL-GNN(Two-Stage) on Cora dataset}
\label{table2:two-stage}
\centering
\setlength{\tabcolsep}{5pt}
\renewcommand{\arraystretch}{0.95}
\begin{tabular}{c|c|c|c}
\toprule
{Dataset} & {Perturbation rate} 
&{Pro-GNN-two}&{RWL(Two-Stage)} 

\\ \midrule

 & 0 

 & 73.31 $\pm$0.71 & {83.62 $\pm$0.821 }\\

& 5 
&73.70 $\pm$1.02 & {78.813 $\pm$ 0.95}\\

{Cora}& 10
 &73.69 $\pm$0.81 & {79.015 $\pm$ 0.843}\\

& 15
&75.38 $\pm$1.10 & { 78.295 $\pm$ 0.862}\\

& 20 
&73.22 $\pm$1.08 & { 78.294 $\pm$ 1.051}\\
& 25
&70.57 $\pm$0.61 & {76.533 $\pm$ 0.72}\\

\bottomrule
\end{tabular}
\end{table}

\subsection{Complexity and Runtime Analysis}
\noindent The proposed framework RWL-GNN has very small computational overhead compared to other proposed defense methods and it can be easily used with the existing GNN architectures like GCN \cite{kipf2017semisupervised}, which have computational complexity of order $\mathcal{O}(|\mathcal{V}|+|\mathcal{E}|)$.The other defense mechanism like ProGNN \cite{jin2020graph}, uses SVD Decomposition to get the low rank adjacency matrix which is costly($\mathcal{O}(|\mathcal{V}|^{3})$,in general). Table \ref{table:time-analysis}, shows a runtime comparison of the different methods for 200 iterations on Cora dataset. We can clearly see that our proposed framework RWL-GNN(Two-Stage) is faster compared with other defense methods except for GNN-Jaccard, and comparing the joint approaches: RWL-GNN(Joint) is faster compared to other joint approaches like Pro-GNN.
\begin{table}[t!] 
\caption{Runtime(in seconds) for 200 epochs of training on cora dataset}
\label{table:time-analysis}
\centering
\renewcommand*{\arraystretch}{1.1}
\begin{tabular}{c|c}
\toprule
{Method} & {Time(s)} \\
\midrule

GCN & 2.17\\
\midrule
GCN-Jaccard & 12.32\\
\midrule
GNNGuard&39.63\\
\midrule
Pro-GNN &220.30\\
\midrule
RWL-GNN(Joint)& 90.90\\
\midrule
RWL-GNN(Two-Stage)&18.55

\\ 
\bottomrule
\end{tabular}
\end{table}

\subsection{Convergence Analysis}
\textbf{Two-Stage Framework Analysis:} We will now show that the limit of the $\b{w}_t$ in \eqref{KKT} satisfies KKT Conditions. The Lagrangian function of the problem \eqref{Final Problem} is : 

\begin{equation*}
    L(\b{w},\b{\mu}) = \mathcal{\alpha} ||\Lw-\Phi_{n}||_{F}^{2} 
  +\mathcal{\beta} Tr(\Lw XX^{T}) - \b{\mu}^{T}\b{w}
\end{equation*}
where $\mu$ is the dual variable.The KKT Conditions of \b{w} for \eqref{Final Problem} and let $\b{w}^{\infty}$ be the limiting point  : 

\begin{align}
 {\mathcal{L}^{*}}{\mathcal{L}}(\b{w}) -{{\mathcal{L}^{*}}}[\Phi_{n} - \frac{\beta}{\alpha} XX^{T}] -\frac{\beta}{\alpha} \mu = 0; \label{m1} \\
  \mu^\top \b{w} =0; \label{m2} \\
\b{w} \geq 0; \label{m3} \\
 \mu \geq 0; \label{m4} 
\end{align}

Also $\b{w}^{\infty}$ is dervied from KKT system of \eqref{KKT}, so we get :

\begin{equation}
\b{w}^{\infty}-(\b{w}^{\infty}-\frac{1}{L_1}({{\mathcal{L}^{*}}}(\Phi_{n} - \frac{\beta}{\alpha} XX^{T})) - \mu_1 = 0
\end{equation}
where $\mu_1$ is the KKT multiplier satisfying the constraints $\mu_1 \geq 0$ and $\mu_1\b{w}^{\infty}=0$. Simplifying the above expression we get : 
\begin{equation}\label{KKT-joint-2}    \mathcal{L}^{*}\mathcal{L}(\b{w}^{\infty})-{{\mathcal{L}^{*}}}(\Phi_{n} - \frac{\beta}{\alpha} XX^{T})-L_1\mu_1 =0 
\end{equation}
Comparing \eqref{m1} and \eqref{KKT-joint-2}, we can see that by making $L_1\mu_1=\frac{\beta}{\alpha}\mu$, the $\b{w}^{\infty}$ also satisfies the KKT Conditions.

    
    

\section{Conclusion}
\noindent In this work we proposed two computationally efficient
frameworks for defending against poisonous attacks on Graph
Neural Networks. The RWL-GNN (Two-Stage) framework cleans the graph
structure first using weighted graph Laplacian, positive semidefiniteness,
feature smoothness properties, and then learn the
model parameters on the clean graph and in the RWL-GNN (joint) framework,
we clean the graph and learn the GNN model parameters jointly. From the set of experiments, we show the
efficacy of our algorithms and also we prove that our proposed
two-stage framework converges to optimal weights. Our
proposed frameworks are easily amenable to other existing
GNN architectures also.

\clearpage
%
%
\bibliographystyle{plainnat}
\bibliography{egbib}
\end{document}